\pdfoutput=1

\PassOptionsToPackage{hyphens}{url}

\documentclass[11pt]{article}

\usepackage{acl}

\usepackage{times}
\usepackage{latexsym}

\usepackage[T1]{fontenc}

\usepackage[utf8]{inputenc}

\usepackage{microtype}

\usepackage{inconsolata}

\usepackage{booktabs}
\usepackage{enumitem}
\usepackage{graphicx}
\usepackage{multirow}

\newcommand{\result}[2]{#1\textsubscript{\,\textcolor{gray}{#2}}}

\title{On the Impact of Calibration Data in\\ Post-training Quantization and Pruning}

\author{Miles Williams \and Nikolaos Aletras \\
  University of Sheffield \\ United Kingdom \\
  \texttt{\{mwilliams15, n.aletras\}@sheffield.ac.uk}}

\begin{document}
\maketitle
\begin{abstract}
Quantization and pruning form the foundation of compression for neural networks, enabling efficient inference for large language models (LLMs). Recently, various quantization and pruning techniques have demonstrated remarkable performance in a post-training setting. They rely upon calibration data, a small set of unlabeled examples that are used to generate layer activations. However, no prior work has systematically investigated how the calibration data impacts the effectiveness of model compression methods. In this paper, we present the first extensive empirical study on the effect of calibration data upon LLM performance. We trial a variety of quantization and pruning methods, datasets, tasks, and models. Surprisingly, we find substantial variations in downstream task performance, contrasting existing work that suggests a greater level of robustness to the calibration data. Finally, we make a series of recommendations for the effective use of calibration data in LLM quantization and pruning.\footnote{\url{https://github.com/mlsw/llm-compression-calibration}}
\end{abstract}

\section{Introduction}

Scaling is an essential component for unlocking new capabilities and improved performance in large language models (LLMs) \citep{brown-etal-2020-language, chowdhery-etal-2023-palm, touvron-etal-2023-llama}. However, the pursuit of scale has led to models that demand significant energy and computational resources \citep{strubell-etal-2019-energy, schwartz-etal-2020-green, wu-etal-2022-sustainable, luccioni-etal-2023-estimating}. Consequently, LLMs can be challenging to deploy, especially in resource-constrained environments \citep{dehghani-etal-2022-efficiency, menghani-2023-efficient}. These challenges have ultimately motivated a substantial body of research on model compression techniques, aiming to reduce computational demands while maintaining performance \citep{treviso-etal-2023-efficient}.

\begin{figure}
\centering
\includegraphics[scale=0.5]{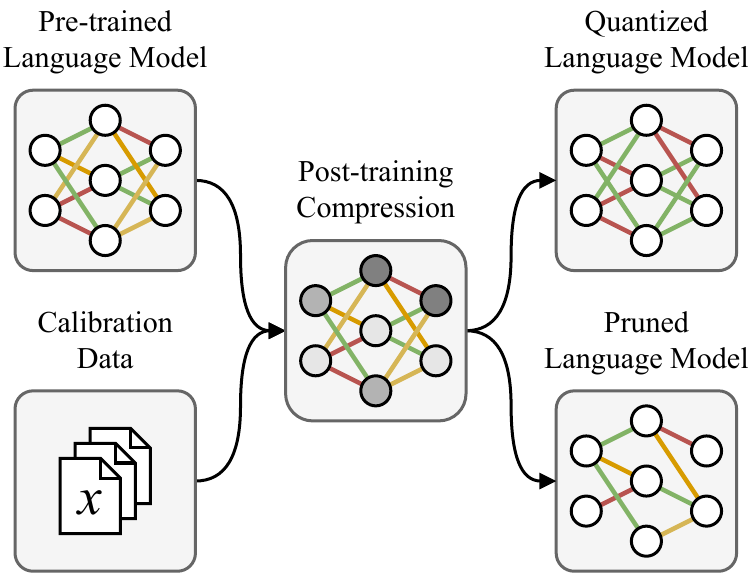}
\caption{Post-training compression methods rely upon calibration data to generate layer activations.}
\label{fig:diagram}
\end{figure}

Quantization and pruning are two of the most popular model compression techniques \citep{gholami-etal-2021-survey, hoefler-etal-2021-sparsity}. Pruning aims to remove redundant weights, while quantization seeks to represent weights (and possibly activations) in lower precision. Most recently, several quantization and pruning methods have demonstrated outstanding performance in a post-training setting \citep{frantar-etal-2023-optq, frantar-alistarh-2023-sparsegpt, dettmers-etal-2024-spqr, sun-etal-2024-simple}.

Post-training compression techniques rely upon \textit{calibration data} \citep{nagel-etal-2020-up, hubara-etal-2021-accurate} to determine the distribution of layer activations. This process requires only a small number of examples, with further examples offering diminishing gains \citep{frantar-alistarh-2023-sparsegpt, sun-etal-2024-simple}. In the case of LLMs, the calibration set is routinely sampled from web text \citep{frantar-etal-2023-optq, frantar-alistarh-2023-sparsegpt, sun-etal-2024-simple} or model pre-training data \citep{xiao-etal-2023-smoothquant, dettmers-etal-2024-spqr}. Notably, the calibration examples are sampled randomly. This is because post-training model compression methods are considered robust to the specific distribution of calibration data \citep{frantar-alistarh-2023-sparsegpt, sun-etal-2024-simple, dettmers-etal-2024-spqr}.

In this paper, we present the first empirical study on the impact of calibration data used in post-training LLM compression. We offer an extensive study with several quantization and pruning methods, across a range of tasks, datasets, and models. Surprisingly, we find that downstream task performance can vary substantially according to the selected calibration data. This contrasts existing work, which suggests a high level of robustness. Finally, we offer a series of recommendations for the effective use of calibration data.

\section{Related Work}
\label{sec:related_work}

\subsection{Model Compression}

Model compression is the process of reducing the memory requirements of a model, usually enabling improved inference efficiency \citep{treviso-etal-2023-efficient}. In the case of neural networks, model compression has a rich history, with origins in seminal work from \citet{lecun-etal-1989-optimal}.\footnote{We refer interested readers to \citet{frantar-alistarh-2022-optimal} for a detailed treatment of quantization and pruning.} Quantization and pruning are two widely adopted approaches for model compression \citep{gholami-etal-2021-survey, hoefler-etal-2021-sparsity}. Pruning seeks to remove redundant weights, while quantization aims to represent the weights (and possibly activations) in lower precision. Applying these techniques to LLMs presents significant challenges, such as large-magnitude outlier features and high computational requirements \citep{dettmers-etal-2022-gpt3, frantar-alistarh-2023-sparsegpt}.

Post-training compression considers the scenario where a model must be compressed without retraining, instead relying upon a small amount of calibration data \citep{nagel-etal-2020-up, hubara-etal-2021-accurate}. While quantization and pruning are distinct methods, they are connected in a post-training setting via the \textit{layer-wise compression problem} \citep{frantar-alistarh-2022-optimal}. This involves selecting compressed weights for each layer that function closely to the original weights, with respect to the calibration data. More formally, given layer $\ell$ with weights $\mathbf{W}_\ell$ and inputs $\mathbf{X}_\ell$, the aim is to minimize $||\mathbf{W}_\ell \mathbf{X}_\ell - \mathbf{\widehat{W}}_\ell \mathbf{X}_\ell||_2^2$ with respect to the compressed weights $\mathbf{\widehat{W}}$. This is subject to a given compression constraint $\mathcal{C}(\mathbf{\widehat{W}}_\ell) > C$, which differs between quantization and pruning.

Recently, a range of approaches have been proposed for the layer-wise compression problem. SparseGPT \citep{frantar-alistarh-2023-sparsegpt} enables LLM pruning up to 60\% sparsity, with little impact upon perplexity. This sequentially prunes the weights in each column of the weight matrix using a series of Hessian inverses, followed by updating the remaining weights. Wanda \citep{sun-etal-2024-simple} significantly improves upon the efficiency of SparseGPT, avoiding the expensive computation of the inverse Hessian. Instead, each layer is pruned according to the product of the weight magnitudes and $\ell_2$ norm of the input activations. For quantization, GPTQ \citep{frantar-etal-2023-optq} enables storing weights in 3- or 4-bit precision, similarly relying upon inverse Hessian information. SpQR \citep{dettmers-etal-2024-spqr} further enables practically lossless quantization of LLMs through identifying and holding outlier weights in higher precision. Other approaches include SmoothQuant \citep{xiao-etal-2023-smoothquant}, which enables 8-bit quantization of weights and activations through shifting the complexity from the activations to the weights.

\subsection{Sampling Calibration Data}

Calibration data is customarily sampled from either web text \citep{frantar-etal-2023-optq, frantar-alistarh-2023-sparsegpt, sun-etal-2024-simple} or model pre-training data \citep{xiao-etal-2023-smoothquant, dettmers-etal-2024-spqr}. The aforementioned approaches use little calibration data, commonly 128 examples, each comprising 2,048 tokens. The addition of further calibration samples provides diminishing gains \citep{frantar-alistarh-2023-sparsegpt, sun-etal-2024-simple}. \citet{frantar-alistarh-2023-sparsegpt} demonstrate that SparseGPT reaches an optimal point with only 128 examples, while \citet{sun-etal-2024-simple} demonstrate that Wanda requires as few as eight examples.

Post-training compression methods are generally understood to be robust to the exact distribution of calibration examples \citep{frantar-alistarh-2023-sparsegpt, sun-etal-2024-simple, dettmers-etal-2024-spqr}. However, concurrent to our own study, there has been a variety of recent work examining the impact of calibration data. \citet{lee-etal-2023-enhancing-computation} note that question-answering performance can vary according to the calibration data sequence length when using GPTQ with 8-bit activation quantization. Moreover, \citet{wu-etal-2023-zeroquant42} find that GPTQ tends to overfit to the calibration data when assessing perplexity on several datasets. Separately, \citet{jaiswal-etal-2024-compressing} suggest that curated calibration data could play an essential role in the design of improved LLM compression methods. Finally, \citet{zeng-etal-2024-multilingual-brain} propose a novel calibration data sampling method to improve multilingual LLM compression.

\subsection{Evaluating Compressed Models}

The recent demand for LLM compression methods has motivated a body of work investigating their broader effects upon model behavior and performance. \citet{goncalves-strubell-2023-understanding} examine the issue of social bias in pre-trained language models, finding that post-training quantization has a regularizing effect. In a separate direction, \citet{chrysostomou-etal-2023-investigating} investigate the impact of pruning upon hallucinations. \citet{kuzmin-etal-2023-pruning} offer the first study that directly compares pruning against quantization. Excluding setups with extreme compression ratios, they find that quantization outperforms pruning. Perhaps the closest work to our own is from \citet{jaiswal-etal-2024-compressing}, which explores the failure modes of compressed LLMs. Notably, they observe that model compression has a substantial impact upon knowledge-intensive tasks.

\section{Methodology}

Our aim is to investigate the robustness of LLM pruning and quantization methods to the calibration data. To this end, we experiment with four compression methods and nine LLMs (i.e. three different sizes from three model families). We vary only the calibration data, trialing five source datasets, each comprising ten distinct calibration sets. This provides a total of 1,800 compressed models. We then evaluate each model across 11 standard NLP tasks, resulting in a total of 19,800 model evaluations.

\subsection{Model Compression}

As it is impractical to exhaustively test every model compression method, we select four popular approaches. Unless otherwise stated, we match the compression setup from the original work. We list complete hyperparameters in Appendix \ref{app:hyperparameters}, Table \ref{tab:hyperparameters}.

\paragraph{Quantization.} For quantization, we consider \textbf{GPTQ} \citep{frantar-etal-2023-optq} and \textbf{SpQR} \citep{dettmers-etal-2024-spqr}. We follow \citet{dettmers-etal-2024-spqr} in using 4-bit weight quantization, which offers an optimal balance between model size and performance \citep{dettmers-etal-2023-case}. Since GPTQ was proposed prior to the release of LLaMA, we use the hyperparameters from the AutoGPTQ library used by Hugging Face Transformers.\footnote{\url{https://github.com/AutoGPTQ/AutoGPTQ}}

\paragraph{Pruning.} For pruning, we trial \textbf{SparseGPT} \citep{frantar-alistarh-2023-sparsegpt} and \textbf{Wanda} \citep{sun-etal-2024-simple}. Since the goal of model compression is typically to enhance inference efficiency, we avoid unstructured pruning, which is challenging to accelerate \citep{wen-etal-2016-learning}. Instead, we opt for semi-structured sparsity, enabling significant inference speedups. We use the 2:4 sparsity pattern that is required for GPU acceleration, resulting in a sparsity ratio of 50\% \citep{mishra-etal-2021-accelerating}.

\subsection{Evaluation Tasks}
\label{sec:evaluation_tasks}

For a fair selection of evaluation tasks, we include all zero-shot tasks used in the original work of GPTQ, SpQR, SparseGPT, and Wanda. These are: (1) ARC easy (ARC-e) and (2) ARC challenge (ARC-c) sets \citep{clark-etal-2018-think}; (3) BoolQ \citep{clark-etal-2019-boolq}; (4) HellaSwag \citep{zellers-etal-2019-hellaswag}; (5) LAMBADA \citep{paperno-etal-2016-lambada}; (6) OpenBookQA \citep{banerjee-etal-2019-careful}; (7) PIQA \citep{bisk-etal-2020-piqa}; (8) RTE \citep{dagan-etal-2006-pascal}; (9) StoryCloze \citep{mostafazadeh-etal-2016-corpus}; and (10) WinoGrande \citep{sakaguchi-etal-2021-winogrande}. 

These zero-shot tasks primarily assess commonsense reasoning abilities, using binary, multiple choice, Cloze, and Winograd style questions. We report the evaluation set sizes in Appendix \ref{app:datasets}.

Finally, following the evaluation protocol used by our selected model compression methods, we also report the model perplexity on the WikiText test set \citep{merity-etal-2017-pointer}.

\subsection{Calibration Data Sources}

We explore a diverse variety of data sources to create our calibration sets. Following previous work (see Section \ref{sec:related_work}), we include random web text and curated model pre-training datasets. To maintain the integrity of the zero-shot evaluations, we avoid using evaluation data as a source of calibration data, following \citet{frantar-etal-2023-optq}. We therefore consider the following five data sources:

\begin{itemize}[left=0pt]
\item \textbf{C4} \citep{raffel-etal-2020-exploring}: We use the Colossal Clean Crawled Corpus as our baseline, following \citet{frantar-etal-2023-optq}. This consists of web text from Common Crawl, filtered with multiple heuristics to form a subset of clean English text.

\item \textbf{CNN-DM} \citep{hermann-etal-2015-teaching, see-etal-2017-get}: The CNN/Daily Mail corpus consists of news articles from both publishers, covering a broad range of topics. We include this corpus since it provides a focused yet distinct genre of high-quality long-form text.

\item \textbf{RedPajama} \citep{together-2023-redpajama}: Since the pre-training data for LLaMA is not publicly available, we instead use an open-source reproduction. This mainly consists of web text (Common Crawl and C4), in addition to selected high-quality sources such as arXiv, GitHub, Stack Exchange, and Books3 \citep{gao-etal-2020-pile}.

\item \textbf{RefinedWeb} \citep{penedo-etal-2023-refinedweb}: Assembled through stringent filtering and deduplication of Common Crawl, RefinedWeb is a curated model pre-training dataset. \citet{penedo-etal-2023-refinedweb} find that models 
trained with this dataset exhibit superior zero-shot generalization abilities compared to alternatives such as The Pile \citep{gao-etal-2020-pile}.

\item \textbf{Wikipedia}: We select English Wikipedia as a source of high-quality encyclopedic text. Specifically, we use a preprocessed and cleaned version of the dump from \texttt{2022-03-01}, prior to the ``knowledge cutoff'' of our selected models.

\end{itemize}

\subsection{Models}

We use three popular `open-source' LLM families: (1) \textbf{LLaMA} \citep{touvron-etal-2023-llama}; (2) \textbf{Vicuna} \citep{chiang-etal-2023-vicuna}; and (3) \textbf{OPT} \citep{zhang-etal-2022-opt}. This includes base models (LLaMA and OPT) as well as instruction-tuned models (Vicuna). 

Additionally, we select these three LLM families since they offer models of comparable sizes. For LLaMA and Vicuna, we select the 7B, 13B, and 33B model sizes. In the case of OPT, we select the closest comparable sizes (6.7B, 13B, and 30B).

\subsection{Implementation Details}

To create the calibration sets, we use the publicly available version of each source dataset from Hugging Face Datasets \citep{lhoest-etal-2021-datasets}. Similarly, we use the weights and implementation of each model from Hugging Face Transformers \citep{wolf-etal-2020-transformers}. To ensure that our model evaluations are robust and reproducible, we use the EleutherAI Language Model Evaluation Harness \citep{gao-etal-20221-framework}. Each model is compressed and evaluated using a single NVIDIA A100 (SXM 80GB) GPU.

\subsection{Data Sampling}

Previous work has demonstrated that increasing the number of calibration examples offers diminishing gains in language modeling performance \citep{frantar-alistarh-2023-sparsegpt, sun-etal-2024-simple}. We therefore follow existing work and randomly sample 128 calibration examples, each consisting of 2,048 tokens \citep{frantar-etal-2023-optq, frantar-alistarh-2023-sparsegpt, sun-etal-2024-simple}. This offers a total of 262,144 tokens in each calibration set. We provide a detailed analysis of how the quantity of calibration examples impacts performance in Section \ref{sec:results}. 

The use of random sampling avoids selection bias and ensures that each calibration set is representative of the source dataset. Similarly, we sample without replacement, to ensure that each calibration example appears only once. To examine the variability introduced by random sampling, we repeat the sampling process to create ten non-overlapping calibration sets for each source dataset. This provides a total of 50 distinct calibration sets.

Due to the vast size of C4, we follow \citet{frantar-etal-2023-optq} in sampling data from the first shard only. We use the same strategy for RefinedWeb, although for RedPajama we use the existing 1B token extract.\footnote{\url{https://huggingface.co/datasets/togethercomputer/RedPajama-Data-1T-Sample}} 

\section{Results \& Analysis}
\label{sec:results}

\begin{figure*}[t]
\centering
\includegraphics{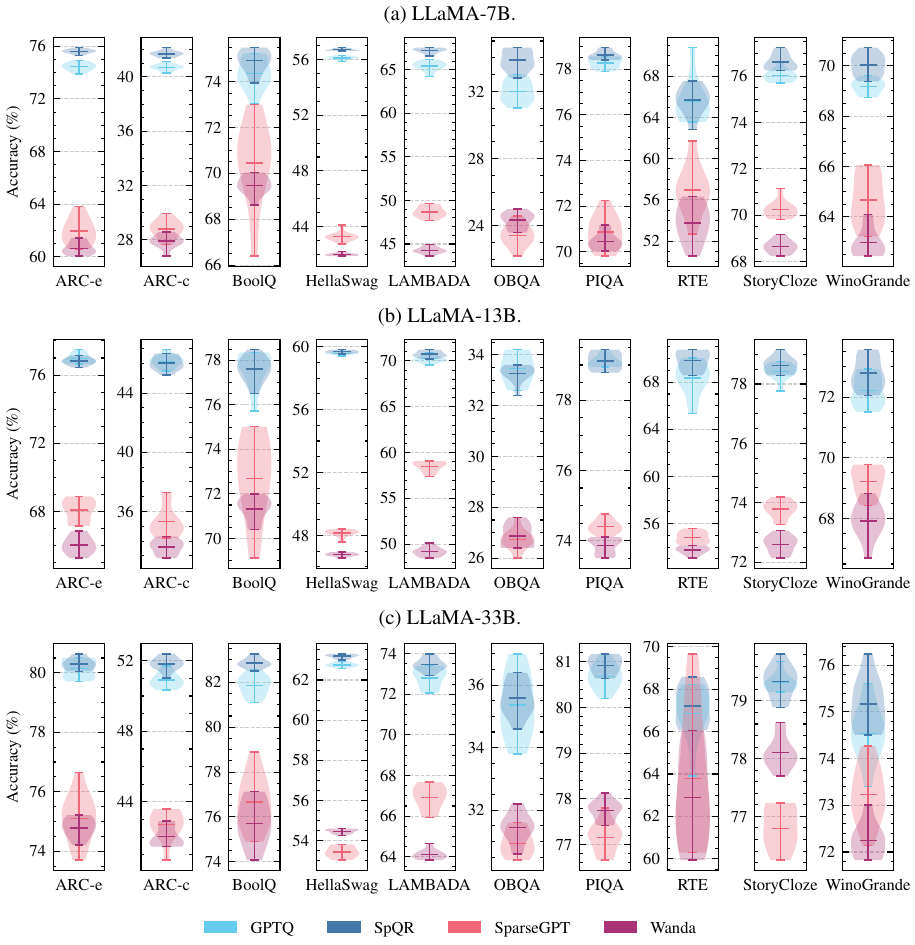}
\caption{Distribution of accuracy across ten calibration sets sampled from C4 for the LLaMA family of models.}
\label{fig:results_tasks_llama}
\end{figure*}

Our principal aim is to investigate the impact of calibration data upon LLM performance. To this end, we examine the model performance across calibration sets, their source datasets, and evaluation tasks. Additionally, we explore trends in overall performance throughout the different configurations.

\paragraph{Performance can vary between calibration sets.} 

Figure \ref{fig:results_tasks_llama} shows the distribution of accuracy across ten calibration sets sampled from the same source dataset (C4) for each model in the LLaMA family. Across several tasks, we observe a substantial level of dispersion. In the most extreme cases with LLaMA-7B, we observe that the accuracy with SparseGPT ranges from 52.7\% to 61.7\% for RTE, and from 66.4\% to 73.0\% for BoolQ. We observe comparable levels of dispersion for the Vicuna and OPT model families, presented in Appendix \ref{app:complete_results} (Figures \ref{fig:results_tasks_vicuna} and \ref{fig:results_tasks_opt}, respectively).

\begin{table*}[t]
\scriptsize
\centering
\begin{tabular}{llrrrrrrrrrr}
\toprule
&  & \multicolumn{3}{c}{LLaMA} & \multicolumn{3}{c}{Vicuna} & \multicolumn{3}{c}{OPT} \\
\cmidrule(lr){3-5} \cmidrule(lr){6-8} \cmidrule(lr){9-11}
Method & Dataset & \multicolumn{1}{c}{7B} & \multicolumn{1}{c}{13B} & \multicolumn{1}{c}{33B} & \multicolumn{1}{c}{7B} & \multicolumn{1}{c}{13B} & \multicolumn{1}{c}{33B} & \multicolumn{1}{c}{6.7B} & \multicolumn{1}{c}{13B} & \multicolumn{1}{c}{30B} \\
\midrule
- & - & \multicolumn{1}{c}{64.3} & \multicolumn{1}{c}{66.7} & \multicolumn{1}{c}{69.1} & \multicolumn{1}{c}{64.3} & \multicolumn{1}{c}{66.5} & \multicolumn{1}{c}{68.8} & \multicolumn{1}{c}{57.4} & \multicolumn{1}{c}{58.3} & \multicolumn{1}{c}{60.4} \\
\cmidrule{1-11}
\multirow[c]{5}{*}{GPTQ} & C4 & \result{63.2}{0.2} & \result{66.2}{0.2} & \result{68.5}{0.2} & \result{62.9}{0.2} & \result{66.1}{0.2} & \result{68.3}{0.2} & \result{56.6}{0.4} & \result{57.7}{0.3} & \result{59.6}{0.2} \\
 & CNN-DM & \result{63.1}{0.3} & \result{66.1}{0.2} & \result{68.5}{0.2} & \result{62.7}{0.3} & \result{66.1}{0.2} & \result{68.4}{0.2} & \result{56.5}{0.3} & \result{57.8}{0.3} & \result{59.6}{0.1} \\
 & RedPajama & \result{63.2}{0.3} & \result{66.2}{0.2} & \result{68.6}{0.2} & \result{63.2}{0.2} & \result{66.2}{0.2} & \result{68.4}{0.2} & \result{56.7}{0.3} & \result{57.9}{0.3} & \result{59.8}{0.3} \\
 & RefinedWeb & \result{63.3}{0.3} & \result{66.1}{0.2} & \result{68.6}{0.2} & \result{63.1}{0.2} & \result{66.0}{0.2} & \result{68.4}{0.2} & \result{56.5}{0.4} & \result{57.8}{0.2} & \result{59.7}{0.2} \\
 & Wikipedia & \result{63.2}{0.2} & \result{66.0}{0.2} & \result{68.6}{0.2} & \result{62.9}{0.3} & \result{66.0}{0.1} & \result{68.1}{0.2} & \result{56.7}{0.3} & \result{57.9}{0.2} & \result{59.7}{0.2} \\
\cmidrule{1-11}
\multirow[c]{5}{*}{SpQR} & C4 & \result{64.1}{0.2} & \result{66.4}{0.1} & \result{69.0}{0.2} & \result{63.9}{0.1} & \result{66.4}{0.2} & \result{68.7}{0.1} & \result{57.1}{0.1} & \result{58.3}{0.1} & \result{60.3}{0.2} \\
 & CNN-DM & \result{63.8}{0.2} & \result{66.5}{0.2} & \result{69.0}{0.1} & \result{64.1}{0.1} & \result{66.3}{0.1} & \result{68.7}{0.2} & \result{57.5}{0.2} & \result{58.3}{0.3} & \result{60.3}{0.1} \\
 & RedPajama & \result{64.0}{0.1} & \result{66.4}{0.2} & \result{68.9}{0.1} & \result{64.1}{0.2} & \result{66.5}{0.2} & \result{68.7}{0.1} & \result{57.2}{0.2} & \result{58.3}{0.2} & \result{60.3}{0.2} \\
 & RefinedWeb & \result{64.1}{0.2} & \result{66.4}{0.2} & \result{69.0}{0.1} & \result{64.0}{0.2} & \result{66.5}{0.2} & \result{68.7}{0.2} & \result{57.1}{0.1} & \result{58.3}{0.1} & \result{60.3}{0.2} \\
 & Wikipedia & \result{63.9}{0.2} & \result{66.5}{0.2} & \result{69.0}{0.1} & \result{64.0}{0.2} & \result{66.3}{0.2} & \result{68.8}{0.1} & \result{57.4}{0.1} & \result{58.3}{0.2} & \result{60.3}{0.2} \\
\cmidrule{1-11}
\multirow[c]{5}{*}{SparseGPT} & C4 & \result{53.9}{0.4} & \result{58.2}{0.3} & \result{63.7}{0.5} & \result{55.7}{0.4} & \result{59.5}{0.4} & \result{64.9}{0.3} & \result{52.9}{0.3} & \result{54.8}{0.3} & \result{57.3}{0.3} \\
 & CNN-DM & \result{52.5}{0.4} & \result{57.5}{0.2} & \result{63.0}{0.4} & \result{52.7}{0.6} & \result{58.9}{0.3} & \result{63.8}{0.3} & \result{51.8}{0.3} & \result{54.0}{0.4} & \result{56.3}{0.2} \\
 & RedPajama & \result{53.3}{0.4} & \result{57.6}{0.2} & \result{63.2}{0.4} & \result{55.5}{0.5} & \result{59.3}{0.4} & \result{64.5}{0.3} & \result{52.6}{0.2} & \result{54.4}{0.3} & \result{56.9}{0.1} \\
 & RefinedWeb & \result{54.0}{0.4} & \result{58.2}{0.2} & \result{63.4}{0.6} & \result{56.3}{0.4} & \result{59.9}{0.3} & \result{65.0}{0.5} & \result{53.0}{0.2} & \result{55.1}{0.1} & \result{57.5}{0.1} \\
 & Wikipedia & \result{52.0}{0.5} & \result{56.3}{0.3} & \result{61.8}{0.4} & \result{54.3}{0.6} & \result{57.6}{0.6} & \result{62.8}{0.3} & \result{51.2}{0.2} & \result{52.7}{0.2} & \result{55.2}{0.2} \\
\cmidrule{1-11}
\multirow[c]{5}{*}{Wanda} & C4 & \result{52.4}{0.3} & \result{56.2}{0.2} & \result{63.4}{0.3} & \result{53.4}{0.4} & \result{56.9}{0.2} & \result{64.0}{0.2} & \result{50.6}{0.2} & \result{52.6}{0.2} & \result{54.6}{0.3} \\
 & CNN-DM & \result{52.4}{0.2} & \result{56.3}{0.1} & \result{63.1}{0.1} & \result{53.7}{0.3} & \result{56.8}{0.2} & \result{63.7}{0.1} & \result{49.6}{0.2} & \result{51.3}{0.4} & \result{54.0}{0.3} \\
 & RedPajama & \result{52.7}{0.2} & \result{56.3}{0.1} & \result{63.1}{0.1} & \result{53.3}{0.2} & \result{57.0}{0.1} & \result{63.8}{0.2} & \result{50.5}{0.1} & \result{52.4}{0.1} & \result{54.8}{0.2} \\
 & RefinedWeb & \result{52.2}{0.3} & \result{56.3}{0.2} & \result{63.2}{0.1} & \result{53.4}{0.3} & \result{57.1}{0.1} & \result{63.7}{0.2} & \result{50.8}{0.1} & \result{52.7}{0.1} & \result{55.1}{0.3} \\
 & Wikipedia & \result{52.4}{0.2} & \result{56.1}{0.2} & \result{62.6}{0.2} & \result{53.4}{0.3} & \result{57.0}{0.1} & \result{63.5}{0.2} & \result{48.9}{0.2} & \result{50.5}{0.2} & \result{53.1}{0.2} \\
\bottomrule
\end{tabular}
\caption{Mean accuracy of all zero-shot tasks across ten calibration sets. Standard deviation is denoted in subscript.}
\label{tab:results_models}
\end{table*}

\paragraph{The degree of dispersion differs between tasks.}

We also observe that the dispersion of accuracy across calibration sets is elevated for certain tasks (Figure \ref{fig:results_tasks_llama}). For example, BoolQ and RTE present the highest levels of dispersion, with ranges up to 6.6\% and 9.4\%, respectively. However, we note that RTE has considerably fewer examples than the other tasks (Appendix \ref{app:datasets}, Table \ref{tab:datasets_statistics}). Consequently, RTE may offer a less stable estimate of dispersion.

\paragraph{Certain data sources may outperform others.}

In the case of pruning, we observe that some calibration data sources achieve greater zero-shot performance for the same model. Table \ref{tab:results_models} shows the mean zero-shot accuracy and standard deviation across ten calibration sets sampled from each source dataset. For SparseGPT, we observe that RefinedWeb reaches the highest level of accuracy for eight of the nine models. In contrast, we observe that Wikipedia achieves the lowest performance across eight of the nine models. Finally, we note that Vicuna-7B demonstrates the greatest range between highest and lowest performing source dataset, with a mean zero-shot accuracy of 52.7\% for CNN-DM versus 56.3\% for RefinedWeb.

\paragraph{No data source consistently outperforms others.}

Although we observe that RefinedWeb generally performs well across various models, this is not universally the case. For example, LLaMA-7B pruned with Wanda achieves the highest mean zero-shot accuracy (52.7\%) with RedPajama, yet the lowest (52.2\%) with RefinedWeb (Table \ref{tab:results_models}). Additionally, we emphasize that the difference between the source datasets with the highest and lowest mean zero-shot accuracy is often very small. Therefore, it is frequently unclear whether a source dataset outperforms its counterparts. This is especially relevant to quantization, where the difference is typically lowest. For example, this is usually around 0.2\% for SpQR, across all model families and sizes.

\paragraph{Calibration data may impact the overall model performance.}

\begin{figure}[t]
\centering
\includegraphics{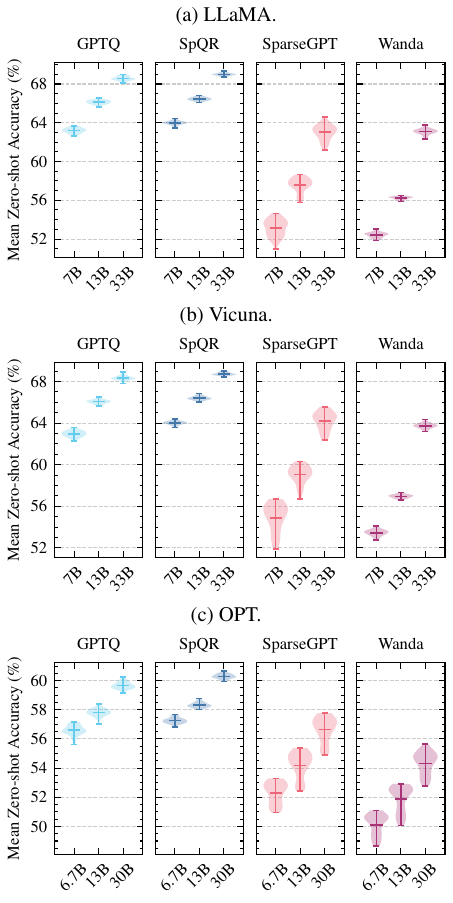}
\caption{The distribution of mean zero-shot accuracy across all calibration sets for every configuration.}
\label{fig:results_models}
\end{figure}

To understand the extent to which calibration data can impact overall model performance, we examine the range between the best and worst performing calibration sets. Figure \ref{fig:results_models} presents the distribution of mean accuracy of all zero-shot tasks across all fifty calibration sets, for each model and compression method. We observe a non-negligible performance range across a variety of models and compression methods. For example, OPT-6.7B has a range of 1.6\% for GPTQ, 0.9\% for SpQR, 2.4\% for SparseGPT, and 2.4\% for Wanda.

\paragraph{Sensitivity can vary between models.}

We observe that the level of sensitivity can differ between models and their families (Figure \ref{fig:results_models}). For pruning, SparseGPT exhibits a greater range (2.4-4.8\%) across models than Wanda (0.6-2.9\%). Interestingly, we also observe that Wanda exhibits a noticeably higher range for the OPT family (0.7-0.9\%) compared to the LLaMA and Vicuna families (0.2-0.3\%). In the case of quantization, we note that GPTQ displays a higher range (0.9-1.6\%) than SpQR (0.6-1.0\%). Again, we observe that the OPT family has a higher range for GPTQ (1.1-1.6\%) than the LLaMA and Vicuna families (0.9-1.3\%).

\paragraph{Quantization methods exhibit lower sensitivity.}

Considering both the individual task (Figures \ref{fig:results_tasks_llama}, \ref{fig:results_tasks_vicuna}, and \ref{fig:results_tasks_opt}) and overall  performance (Figure \ref{fig:results_models}), we observe that the quantization methods generally exhibit lower levels of dispersion than the pruning methods. Notably, the practically lossless compression offered by SpQR appears to leave little room for sensitivity to the calibration data. This suggests that while the exact distribution of calibration data may impact model performance, the compression method itself may be a more influential factor.

\paragraph{Pruning methods exhibit higher sensitivity.}

Contrasting the quantization methods, we observe that the pruning methods generally display higher dispersion. In particular, SparseGPT consistently demonstrates the highest levels of dispersion (Figure \ref{fig:results_models}). We speculate that this is partly due to the destructive nature of pruning, in comparison to quantization. For example, Table \ref{tab:results_models} shows that the quantized models offer comparable performance to the original models, whereas there is a substantial performance degradation for the pruned models.

\begin{figure*}[t]
\begin{minipage}{\columnwidth}
\centering
\includegraphics{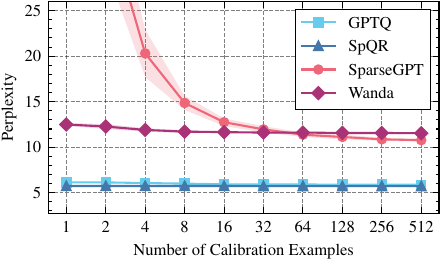}
\end{minipage}%
\hspace{\columnsep}%
\begin{minipage}{\columnwidth}
\centering
\includegraphics{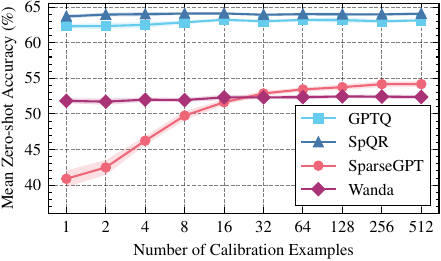}
\end{minipage}
\caption{The perplexity on WikiText (L) and mean zero-shot accuracy (R) for LLaMA-7B with each compression method. We present the mean value and standard deviation (shaded) across ten calibration sets sampled from C4.}
\label{fig:results_quantity_llama}
\end{figure*}

\begin{table*}[t]
\scriptsize
\centering
\begin{tabular}{llrrrrrrrrr}
\toprule
&  & \multicolumn{3}{c}{LLaMA} & \multicolumn{3}{c}{Vicuna} & \multicolumn{3}{c}{OPT} \\
\cmidrule(lr){3-5} \cmidrule(lr){6-8} \cmidrule(lr){9-11}
Method & Dataset & \multicolumn{1}{c}{7B} & \multicolumn{1}{c}{13B} & \multicolumn{1}{c}{33B} & \multicolumn{1}{c}{7B} & \multicolumn{1}{c}{13B} & \multicolumn{1}{c}{33B} & \multicolumn{1}{c}{6.7B} & \multicolumn{1}{c}{13B} & \multicolumn{1}{c}{30B} \\
\midrule
- & - & \multicolumn{1}{c}{5.68} & \multicolumn{1}{c}{5.09} & \multicolumn{1}{c}{4.10} & \multicolumn{1}{c}{6.90} & \multicolumn{1}{c}{6.09} & \multicolumn{1}{c}{5.22} & \multicolumn{1}{c}{10.86} & \multicolumn{1}{c}{10.13} & \multicolumn{1}{c}{9.56} \\
\cmidrule{1-11}
\multirow[c]{5}{*}{GPTQ} & C4 & \result{5.90}{0.02} & \result{5.22}{0.01} & \result{4.27}{0.01} & \result{7.10}{0.02} & \result{6.21}{0.02} & \result{5.42}{0.02} & \result{10.93}{0.08} & \result{10.27}{0.03} & \result{9.58}{0.05} \\
 & CNN-DM & \result{5.90}{0.02} & \result{5.34}{0.01} & \result{4.29}{0.01} & \result{7.12}{0.02} & \result{6.37}{0.02} & \result{5.45}{0.03} & \result{10.99}{0.04} & \result{10.27}{0.03} & \result{9.58}{0.02} \\
 & RedPajama & \result{5.91}{0.02} & \result{5.22}{0.01} & \result{4.27}{0.01} & \result{7.10}{0.04} & \result{6.22}{0.02} & \result{5.43}{0.02} & \result{10.99}{0.05} & \result{10.27}{0.03} & \result{9.55}{0.07} \\
 & RefinedWeb & \result{5.91}{0.02} & \result{5.22}{0.01} & \result{4.27}{0.01} & \result{7.05}{0.03} & \result{6.23}{0.01} & \result{5.43}{0.02} & \result{10.99}{0.06} & \result{10.25}{0.04} & \result{9.57}{0.03} \\
 & Wikipedia & \result{5.85}{0.01} & \result{5.21}{0.01} & \result{4.24}{0.01} & \result{7.07}{0.03} & \result{6.19}{0.01} & \result{5.41}{0.02} & \result{10.95}{0.09} & \result{10.26}{0.04} & \result{9.57}{0.05} \\
\cmidrule{1-11}
\multirow[c]{5}{*}{SpQR} & C4 & \result{5.74}{0.01} & \result{5.13}{0.01} & \result{4.15}{0.01} & \result{6.95}{0.01} & \result{6.12}{0.01} & \result{5.28}{0.01} & \result{10.88}{0.03} & \result{10.16}{0.02} & \result{9.46}{0.05} \\
 & CNN-DM & \result{5.74}{0.01} & \result{5.14}{0.01} & \result{4.15}{0.00} & \result{6.93}{0.02} & \result{6.12}{0.01} & \result{5.29}{0.01} & \result{10.83}{0.09} & \result{10.24}{0.02} & \result{9.46}{0.04} \\
 & RedPajama & \result{5.73}{0.00} & \result{5.13}{0.01} & \result{4.15}{0.00} & \result{6.93}{0.02} & \result{6.12}{0.01} & \result{5.28}{0.01} & \result{10.90}{0.02} & \result{10.16}{0.02} & \result{9.45}{0.04} \\
 & RefinedWeb & \result{5.74}{0.01} & \result{5.13}{0.01} & \result{4.15}{0.00} & \result{6.94}{0.02} & \result{6.11}{0.01} & \result{5.29}{0.01} & \result{10.90}{0.02} & \result{10.14}{0.02} & \result{9.47}{0.03} \\
 & Wikipedia & \result{5.73}{0.01} & \result{5.13}{0.01} & \result{4.14}{0.00} & \result{6.93}{0.02} & \result{6.11}{0.01} & \result{5.28}{0.01} & \result{10.89}{0.02} & \result{10.18}{0.01} & \result{9.46}{0.05} \\
\cmidrule{1-11}
\multirow[c]{5}{*}{SparseGPT} & C4 & \result{11.06}{0.18} & \result{9.11}{0.10} & \result{7.21}{0.08} & \result{12.39}{0.31} & \result{9.92}{0.14} & \result{8.14}{0.07} & \result{14.25}{0.13} & \result{12.93}{0.07} & \result{10.92}{0.09} \\
 & CNN-DM & \result{11.32}{0.17} & \result{8.42}{0.05} & \result{6.78}{0.04} & \result{12.72}{0.18} & \result{9.28}{0.05} & \result{7.80}{0.05} & \result{14.55}{0.12} & \result{13.46}{0.13} & \result{11.13}{0.05} \\
 & RedPajama & \result{10.71}{0.17} & \result{8.82}{0.09} & \result{7.03}{0.07} & \result{12.04}{0.13} & \result{9.49}{0.10} & \result{8.06}{0.07} & \result{14.15}{0.11} & \result{13.02}{0.11} & \result{11.18}{0.06} \\
 & RefinedWeb & \result{10.91}{0.13} & \result{8.99}{0.07} & \result{7.14}{0.07} & \result{12.27}{0.19} & \result{9.70}{0.09} & \result{8.12}{0.07} & \result{13.97}{0.13} & \result{12.67}{0.07} & \result{10.73}{0.07} \\
 & Wikipedia & \result{9.96}{0.15} & \result{8.33}{0.10} & \result{6.75}{0.08} & \result{11.28}{0.15} & \result{9.04}{0.11} & \result{7.73}{0.07} & \result{14.46}{0.12} & \result{13.33}{0.14} & \result{11.62}{0.05} \\
\cmidrule{1-11}
\multirow[c]{5}{*}{Wanda} & C4 & \result{11.57}{0.05} & \result{9.70}{0.04} & \result{6.97}{0.02} & \result{13.85}{0.10} & \result{10.99}{0.06} & \result{8.51}{0.04} & \result{16.04}{0.17} & \result{15.64}{0.13} & \result{13.39}{0.21} \\
 & CNN-DM & \result{11.31}{0.04} & \result{9.39}{0.04} & \result{6.83}{0.01} & \result{13.44}{0.06} & \result{10.63}{0.05} & \result{8.38}{0.02} & \result{16.60}{0.09} & \result{17.29}{0.08} & \result{13.50}{0.22} \\
 & RedPajama & \result{11.38}{0.05} & \result{9.45}{0.05} & \result{6.90}{0.02} & \result{13.51}{0.14} & \result{10.62}{0.04} & \result{8.45}{0.02} & \result{16.22}{0.11} & \result{15.99}{0.18} & \result{13.63}{0.20} \\
 & RefinedWeb & \result{11.55}{0.05} & \result{9.56}{0.05} & \result{6.92}{0.01} & \result{13.76}{0.11} & \result{10.76}{0.06} & \result{8.47}{0.02} & \result{15.76}{0.12} & \result{15.34}{0.11} & \result{13.13}{0.41} \\
 & Wikipedia & \result{11.04}{0.03} & \result{9.31}{0.03} & \result{6.84}{0.02} & \result{13.09}{0.04} & \result{10.53}{0.05} & \result{8.38}{0.04} & \result{16.21}{0.10} & \result{16.32}{0.10} & \result{13.76}{0.23} \\
\bottomrule
\end{tabular}
\caption{Mean perplexity on WikiText across ten calibration sets. Standard deviation is denoted in subscript.}
\label{tab:results_perplexity}
\end{table*}

\paragraph{Additional calibration examples offer diminishing gains in language modeling performance.}

Figure \ref{fig:results_quantity_llama} presents the perplexity on WikiText for LLaMA-7B compressed with increasing quantities of calibration data sampled from C4. For quantization, GPTQ improves from 6.13$\pm$0.05 with a single example to 5.90$\pm$0.03 with 128, while SpQR remains almost constant at 5.74$\pm$0.01. In the case of pruning, SparseGPT gradually improves from 55.26$\pm$10.82 with one example to 11.12$\pm$0.26 with 128, while Wanda improves from 12.50$\pm$0.15 to 11.56$\pm$0.06. This corroborates findings from previous work that only a small number of examples are required to maximize language modeling performance \citep{frantar-etal-2023-optq, sun-etal-2024-simple}.

\paragraph{Additional calibration examples offer diminishing gains in zero-shot performance.}

Figure \ref{fig:results_quantity_llama} also presents the mean zero-shot accuracy across tasks (Section \ref{sec:evaluation_tasks}) for LLaMA-7B compressed using an increasing number of calibration examples. For each compression method, we observe an identical trend to perplexity, with performance plateauing after only a small number of examples. Remarkably, GPTQ, SpQR, and Wanda demonstrate a consistent mean zero-shot accuracy with only a few examples. However, SparseGPT performance continues to improve beyond 128 examples, with quadruple the number of examples offering a 0.4\% increase in mean zero-shot accuracy. This suggests that SparseGPT is best used with an expanded calibration set to maximize performance. However, it should be noted that increasing the number of calibration examples also increases the computational cost of compression \citep{frantar-alistarh-2023-sparsegpt}.

\paragraph{Perplexity can be challenging to interpret.}

Prior work relies on perplexity to assess robustness to the calibration data \citep{frantar-alistarh-2023-sparsegpt, sun-etal-2024-simple}. In practice, this consists of compressing a given model using a few different calibration sets, sampled from the same source dataset. Then, the difference in perplexity is measured using a standard dataset (often WikiText). Although this is an entirely reasonable approach, it can be challenging to interpret these results in the context of downstream task performance. Table \ref{tab:results_perplexity}, shows the WikiText test set perplexity and standard deviation across ten distinct calibration sets sampled from each source dataset. For example, applying SparseGPT to Vicuna-7B with the calibration sets from CNN-DM achieves a seemingly robust perplexity of 12.72$\pm$0.18. In contrast, the same models achieve 66.7\%$\pm$4.7 on BoolQ, with accuracy ranging from 57.0\% to 71.6\%.

\paragraph{SparseGPT typically outperforms Wanda.}

Although the purpose of our study is not to compare model compression methods, we observe that SparseGPT mostly outperforms Wanda. Considering the C4 calibration data source that was used in both original works, we observe that SparseGPT achieves a higher mean zero-shot accuracy across all models. Moreover, we observe that the mean zero-shot accuracy achieved by SparseGPT is 2.2-2.6\% higher across the OPT family, compared to 0.3-2.0\% and 0.9-2.6\% for LLaMA and Vicuna, respectively (Table \ref{tab:results_models}). Although \citet{sun-etal-2024-simple} report mixed results for Wanda versus SparseGPT in the 2:4 semi-structured setting, we were somewhat surprised by the consistency and margin that SparseGPT outperforms Wanda in our experiments.

\section{Recommendations}

Our results suggest that calibration data used in post-training quantization and pruning can influence LLM performance. Consequently, we offer several recommendations concerning the use of calibration data to researchers and practitioners alike: 

\begin{enumerate}[left=0pt]
\item \textbf{Releasing calibration data}:
Research relying upon calibration data for post-training model compression should release the data. As calibration data can ultimately affect model performance, this serves to improve reproducibility by removing a source of randomness.\footnote{Many existing studies have released code to generate the calibration data, but not the calibration data itself. This is not robust since using a seed for random sampling does not guarantee consistency across software or platform versions.}

\item \textbf{Varying calibration data}:
Evaluating downstream task performance across several calibration sets during model development can offer an insight into the sensitivity to the calibration data. This provides an opportunity to identify any issues with the compression setup or calibration data that may compromise model performance.

\item \textbf{Inspecting calibration data}:
Randomly sampled calibration data can be manually inspected, to remove anomalous examples. Given the small number of examples comprising the calibration set, this may offer a practical way to maximize the performance of the compressed model.
\end{enumerate}

\section{Conclusion}

In this paper, we presented the first extensive empirical study on calibration data for LLM quantization and pruning. We examined the downstream task performance across a variety of models, compression methods, and calibration data sources. Our results suggest that calibration data can substantially influence the performance of compressed LLMs. We supplement our findings with several recommendations for the effective use of calibration data.

We hope that our work will inspire further research surrounding the use of calibration data in LLM compression, an area which has seen limited attention. For future work, we are particularly interested in exploring how choices in the training protocol may influence the sensitivity to calibration data in LLM compression \citep{ahmadian-etal-2023-intriguing}.

\section*{Limitations}

In this study, we rely entirely on English-language models, evaluation tasks, and calibration data. We operate under the assumption that the performance of the LLM compression methods we trial is generally language-agnostic. Nevertheless, we recognize the importance of linguistic diversity. We therefore hope to explore the performance of LLM compression methods across diverse language families (including low resource settings) in a future work.

\section*{Acknowledgments}

We are grateful to George Chrysostomou, Huiyin Xue, and the anonymous reviewers for their invaluable feedback. MW is supported by the Centre for Doctoral Training in Speech and Language Technologies (SLT) and their Applications funded by UK Research and Innovation grant EP/S023062/1. NA is supported by EPSRC grant EP/Y009800/1, part of the RAI UK Keystone projects.

\bibliography{anthology,custom}

\appendix

\section{Hyperparameters}
\label{app:hyperparameters}

Table \ref{tab:hyperparameters} lists the complete hyperparameters used for each compression method. For SparseGPT, Wanda, and SpQR, these are taken from the original work. In the case of GPTQ, which was proposed prior to LLaMA, these are derived from AutoGPTQ.

\begin{table}[th]
\small
\centering
\begin{tabular}{llc}
\toprule
Method & Hyperparameter & Value \\
\midrule
\multirow{6}{*}{GPTQ}  & Bits per Weight & 4 \\
 & Dampening & 0.01 \\
 & Descending Activation Order & Yes \\
 & Group Size & 128 \\
 & Symmetric Quantization & Yes \\
 & True Sequential Quantization & Yes \\
\midrule
\multirow{10}{*}{SpQR}  & Bits per Scale & 3 \\
 & Bits per Weight & 4 \\
 & Bits per Zero & 3 \\
 & Dampening & 1.0 \\
 & Descending Activation Order & Yes \\
 & Group Size (Weights) & 16 \\
 & Group Size (Statistics) & 16 \\
 & Symmetric Quantization & No \\
 & True Sequential Quantization & No \\
 & Outlier Threshold & 0.2 \\
 \midrule
 \multirow{3}{*}{SparseGPT} & Dampening & 0.01 \\
 & Group Size & 128 \\
 & Sparsity & 2:4 \\
\midrule
\multirow{2}{*}{Wanda} & Group Size & 1 \\
 & Sparsity & 2:4 \\
\bottomrule
\end{tabular}
\caption{The hyperparameters used for all experiments.}
\label{tab:hyperparameters}
\end{table}

\section{Datasets}
\label{app:datasets}

Table \ref{tab:datasets_statistics} shows the number of examples for each evaluation task from the appropriate split (either test or validation).

\begin{table}[th]
\small
\centering
\begin{tabular}{lr}
\toprule
Dataset & \#
Examples \\
\midrule
ARC-Easy \citep{clark-etal-2018-think} & 2,376 \\
ARC-Challenge \citep{clark-etal-2018-think} & 1,172 \\
BoolQ \citep{clark-etal-2019-boolq} & 3,270 \\
HellaSwag \citep{zellers-etal-2019-hellaswag} & 10,042 \\
LAMBADA \citep{paperno-etal-2016-lambada} & 5,153 \\
OpenBookQA \citep{banerjee-etal-2019-careful} & 500 \\
PIQA \citep{bisk-etal-2020-piqa} & 1,838 \\
RTE \citep{dagan-etal-2006-pascal} & 277 \\
StoryCloze \citep{mostafazadeh-etal-2016-corpus} & 1,511 \\
WinoGrande \citep{sakaguchi-etal-2021-winogrande} & 1,267 \\
\bottomrule
\end{tabular}
\caption{Number of examples for each evaluation task.}
\label{tab:datasets_statistics}
\end{table}

\section{Complete Results}
\label{app:complete_results}

\begin{figure*}[t]
\centering
\includegraphics{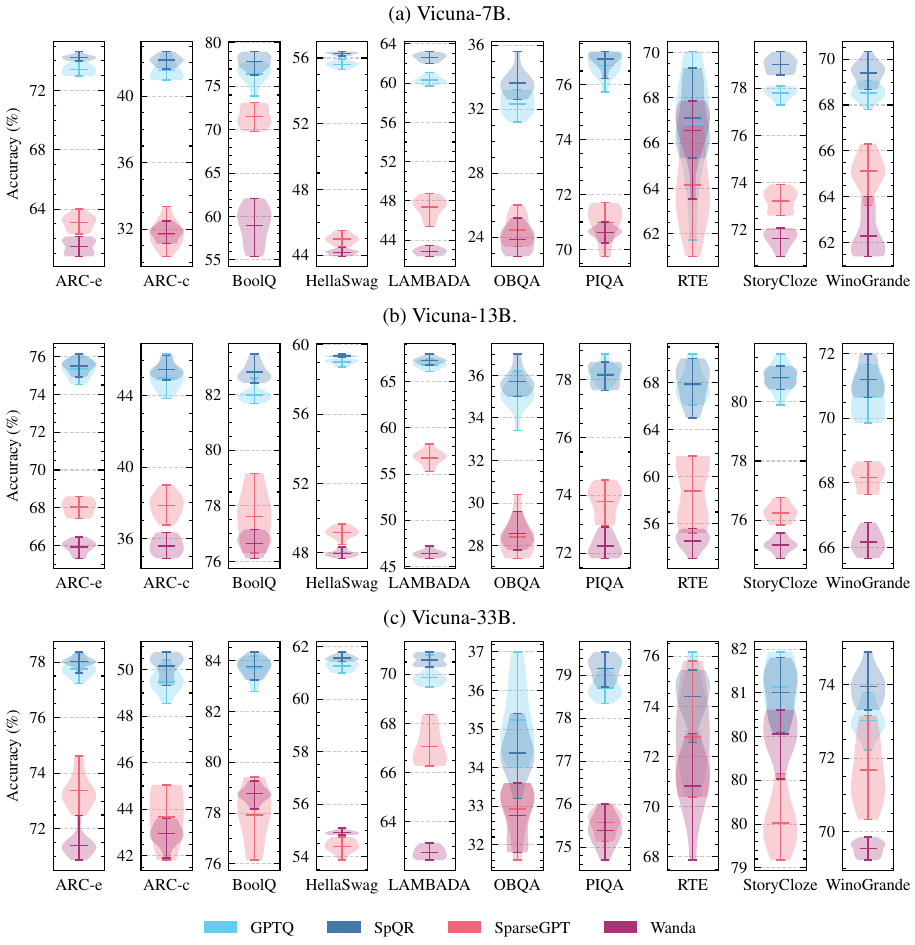}
\caption{Distribution of accuracy across ten calibration sets sampled from C4 for the Vicuna family of models.}
\label{fig:results_tasks_vicuna}
\end{figure*}

\begin{figure*}[t]
\centering
\includegraphics{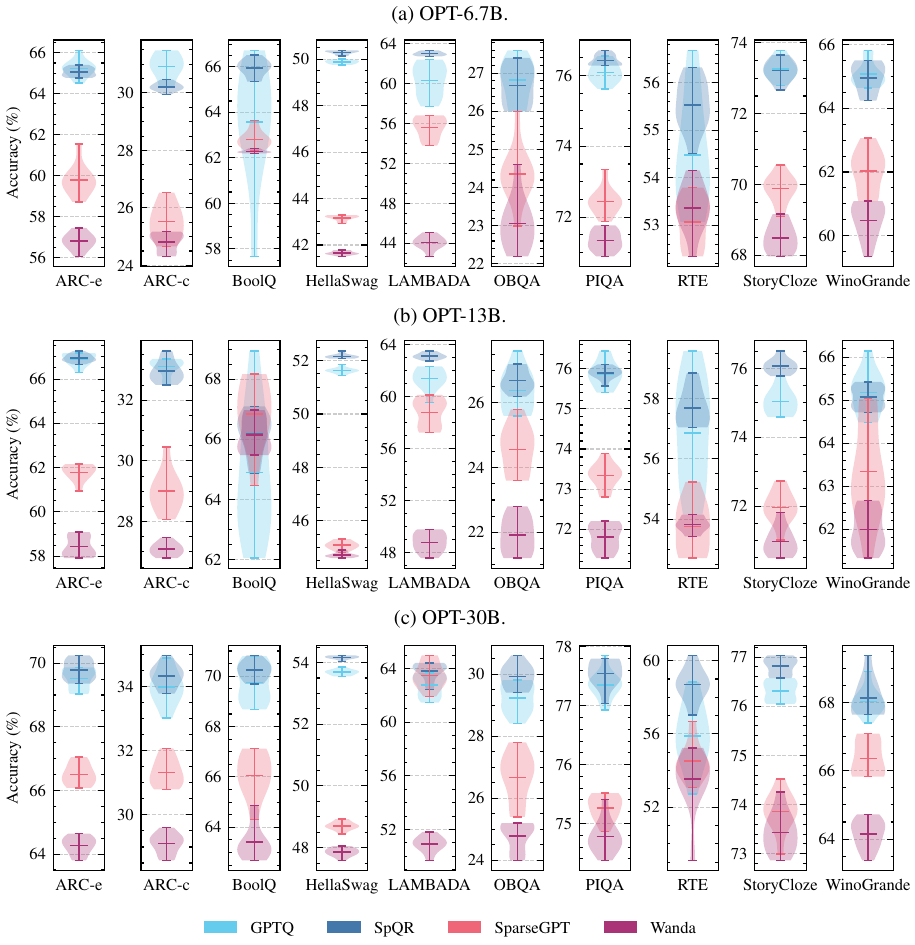}
\caption{Distribution of accuracy across ten calibration sets sampled from C4 for the OPT family of models.}
\label{fig:results_tasks_opt}
\end{figure*}

In addition to the summarized results (Section \ref{sec:results}), we provide a complete tabulation of results for all models, tasks, compression methods, and source datasets. We present the mean accuracy and standard deviation across ten distinct calibration sets for each model family: LLaMA 7B, 13B, and 33B (Tables \ref{tab:results_tasks_llama_7b}, \ref{tab:results_tasks_llama_13b} and \ref{tab:results_tasks_llama_33b}); Vicuna 7B, 13B, and 33B (Tables \ref{tab:results_tasks_vicuna_7b_v1.3}, \ref{tab:results_tasks_vicuna_13b_v1.3}, and \ref{tab:results_tasks_vicuna_33b_v1.3}); and OPT 6.7B, 13B, and 30B (Tables \ref{tab:results_tasks_opt_6.7b}, \ref{tab:results_tasks_opt_13b}, and \ref{tab:results_tasks_opt_30b}). Finally, we present the distribution of accuracy across calibration sets sampled from C4 (Figure \ref{fig:results_tasks_llama}) for the Vicuna and OPT model families in Figures \ref{fig:results_tasks_opt} and \ref{fig:results_tasks_vicuna}, respectively.

\begin{table*}[t]
\scriptsize
\centering
\begin{tabular}{llrrrrrrrrrrr}
\toprule
Method & Dataset & \multicolumn{1}{c}{ARC-e} & \multicolumn{1}{c}{ARC-c} & \multicolumn{1}{c}{BoolQ} & \multicolumn{1}{c}{HS} & \multicolumn{1}{c}{LMBD} & \multicolumn{1}{c}{OBQA} & \multicolumn{1}{c}{PIQA} & \multicolumn{1}{c}{RTE} & \multicolumn{1}{c}{SC} & \multicolumn{1}{c}{WG} & \multicolumn{1}{c}{Mean} \\
\midrule
- & - & \multicolumn{1}{c}{75.3} & \multicolumn{1}{c}{41.9} & \multicolumn{1}{c}{75.1} & \multicolumn{1}{c}{57.0} & \multicolumn{1}{c}{67.8} & \multicolumn{1}{c}{34.4} & \multicolumn{1}{c}{78.7} & \multicolumn{1}{c}{66.4} & \multicolumn{1}{c}{76.7} & \multicolumn{1}{c}{69.9} & \multicolumn{1}{c}{64.3} \\
\cmidrule{1-13}
\multirow[c]{5}{*}{GPTQ} & C4 & \result{74.4}{0.3} & \result{40.7}{0.2} & \result{74.4}{0.7} & \result{56.1}{0.1} & \result{65.4}{0.6} & \result{32.0}{0.7} & \result{78.3}{0.2} & \result{65.6}{2.1} & \result{76.0}{0.3} & \result{69.2}{0.3} & \result{63.2}{0.2} \\
 & CNN-DM & \result{74.3}{0.3} & \result{40.9}{0.5} & \result{74.1}{0.5} & \result{56.1}{0.1} & \result{64.7}{0.6} & \result{32.1}{0.9} & \result{78.3}{0.3} & \result{65.7}{1.3} & \result{75.8}{0.2} & \result{69.2}{0.8} & \result{63.1}{0.3} \\
 & RedPajama & \result{74.1}{0.4} & \result{40.5}{0.7} & \result{74.0}{0.5} & \result{56.2}{0.1} & \result{65.3}{0.6} & \result{31.7}{0.8} & \result{78.4}{0.2} & \result{66.5}{2.6} & \result{76.0}{0.4} & \result{68.8}{0.5} & \result{63.2}{0.3} \\
 & RefinedWeb & \result{74.0}{0.5} & \result{40.9}{0.6} & \result{74.6}{0.6} & \result{56.3}{0.1} & \result{65.6}{0.4} & \result{32.0}{1.0} & \result{78.4}{0.3} & \result{65.4}{2.2} & \result{76.0}{0.2} & \result{69.5}{0.4} & \result{63.3}{0.3} \\
 & Wikipedia & \result{74.4}{0.3} & \result{41.4}{0.9} & \result{74.4}{0.5} & \result{56.3}{0.2} & \result{65.0}{0.4} & \result{31.7}{0.8} & \result{78.5}{0.2} & \result{64.6}{2.3} & \result{76.2}{0.4} & \result{69.5}{0.4} & \result{63.2}{0.2} \\
\bottomrule
\end{tabular}
\caption{Mean accuracy across ten calibration sets for LLaMA-7B, with standard deviation denoted in subscript.}
\label{tab:results_tasks_llama_7b}
\end{table*}

\begin{table*}[t]
\scriptsize
\centering
\begin{tabular}{llrrrrrrrrrrr}
\toprule
Method & Dataset & \multicolumn{1}{c}{ARC-e} & \multicolumn{1}{c}{ARC-c} & \multicolumn{1}{c}{BoolQ} & \multicolumn{1}{c}{HS} & \multicolumn{1}{c}{LMBD} & \multicolumn{1}{c}{OBQA} & \multicolumn{1}{c}{PIQA} & \multicolumn{1}{c}{RTE} & \multicolumn{1}{c}{SC} & \multicolumn{1}{c}{WG} & \multicolumn{1}{c}{Mean} \\
\midrule
- & - & \multicolumn{1}{c}{77.4} & \multicolumn{1}{c}{46.5} & \multicolumn{1}{c}{78.0} & \multicolumn{1}{c}{59.9} & \multicolumn{1}{c}{71.1} & \multicolumn{1}{c}{33.2} & \multicolumn{1}{c}{79.2} & \multicolumn{1}{c}{70.4} & \multicolumn{1}{c}{78.5} & \multicolumn{1}{c}{72.8} & \multicolumn{1}{c}{66.7} \\
\cmidrule{1-13}
\multirow[c]{5}{*}{GPTQ} & C4 & \result{77.0}{0.3} & \result{46.1}{0.4} & \result{77.6}{0.8} & \result{59.6}{0.1} & \result{70.3}{0.4} & \result{33.5}{0.6} & \result{79.1}{0.2} & \result{68.3}{1.5} & \result{78.4}{0.3} & \result{72.2}{0.5} & \result{66.2}{0.2} \\
 & CNN-DM & \result{76.9}{0.4} & \result{46.0}{0.7} & \result{77.8}{0.4} & \result{59.4}{0.2} & \result{70.4}{0.3} & \result{33.1}{0.6} & \result{78.9}{0.2} & \result{68.0}{2.0} & \result{78.4}{0.3} & \result{72.4}{0.4} & \result{66.1}{0.2} \\
 & RedPajama & \result{77.2}{0.4} & \result{46.2}{0.7} & \result{77.4}{0.6} & \result{59.6}{0.2} & \result{70.3}{0.2} & \result{33.5}{0.6} & \result{79.1}{0.2} & \result{68.0}{1.9} & \result{78.4}{0.2} & \result{72.4}{0.5} & \result{66.2}{0.2} \\
 & RefinedWeb & \result{77.0}{0.4} & \result{46.2}{0.4} & \result{77.6}{0.6} & \result{59.5}{0.1} & \result{70.4}{0.4} & \result{33.6}{0.3} & \result{79.2}{0.3} & \result{67.3}{1.5} & \result{78.3}{0.3} & \result{72.4}{0.4} & \result{66.1}{0.2} \\
 & Wikipedia & \result{76.9}{0.4} & \result{46.1}{0.4} & \result{77.0}{0.7} & \result{59.4}{0.2} & \result{70.4}{0.6} & \result{33.4}{0.5} & \result{79.0}{0.3} & \result{67.4}{1.8} & \result{78.1}{0.3} & \result{72.2}{0.5} & \result{66.0}{0.2} \\
\bottomrule
\end{tabular}
\caption{Mean accuracy across ten calibration sets for LLaMA-13B, with standard deviation denoted in subscript.}
\label{tab:results_tasks_llama_13b}
\end{table*}

\begin{table*}[t]
\scriptsize
\centering
\begin{tabular}{llrrrrrrrrrrr}
\toprule
Method & Dataset & \multicolumn{1}{c}{ARC-e} & \multicolumn{1}{c}{ARC-c} & \multicolumn{1}{c}{BoolQ} & \multicolumn{1}{c}{HS} & \multicolumn{1}{c}{LMBD} & \multicolumn{1}{c}{OBQA} & \multicolumn{1}{c}{PIQA} & \multicolumn{1}{c}{RTE} & \multicolumn{1}{c}{SC} & \multicolumn{1}{c}{WG} & \multicolumn{1}{c}{Mean} \\
\midrule
- & - & \multicolumn{1}{c}{80.4} & \multicolumn{1}{c}{52.8} & \multicolumn{1}{c}{82.6} & \multicolumn{1}{c}{63.3} & \multicolumn{1}{c}{73.4} & \multicolumn{1}{c}{36.0} & \multicolumn{1}{c}{81.0} & \multicolumn{1}{c}{66.8} & \multicolumn{1}{c}{79.1} & \multicolumn{1}{c}{75.9} & \multicolumn{1}{c}{69.1} \\
\cmidrule{1-13}
\multirow[c]{5}{*}{GPTQ} & C4 & \result{80.1}{0.3} & \result{50.9}{0.4} & \result{81.9}{0.5} & \result{62.8}{0.1} & \result{72.8}{0.5} & \result{35.4}{1.0} & \result{80.7}{0.3} & \result{66.9}{1.3} & \result{79.3}{0.2} & \result{74.5}{0.6} & \result{68.5}{0.2} \\
 & CNN-DM & \result{79.7}{0.5} & \result{50.4}{0.4} & \result{82.2}{0.7} & \result{62.6}{0.1} & \result{73.0}{0.4} & \result{35.2}{0.7} & \result{80.7}{0.2} & \result{67.2}{1.9} & \result{79.2}{0.3} & \result{75.2}{0.4} & \result{68.5}{0.2} \\
 & RedPajama & \result{79.9}{0.4} & \result{50.7}{0.8} & \result{81.7}{0.7} & \result{62.7}{0.1} & \result{73.1}{0.2} & \result{35.2}{0.5} & \result{80.6}{0.2} & \result{68.0}{1.5} & \result{79.1}{0.2} & \result{74.9}{0.4} & \result{68.6}{0.2} \\
 & RefinedWeb & \result{79.5}{0.3} & \result{50.7}{0.4} & \result{82.3}{0.4} & \result{62.7}{0.1} & \result{72.9}{0.5} & \result{35.2}{0.7} & \result{80.8}{0.3} & \result{68.1}{1.5} & \result{79.3}{0.2} & \result{74.9}{0.3} & \result{68.6}{0.2} \\
 & Wikipedia & \result{80.1}{0.4} & \result{50.6}{0.7} & \result{82.0}{0.7} & \result{62.7}{0.1} & \result{73.0}{0.4} & \result{35.0}{0.6} & \result{80.7}{0.3} & \result{67.4}{1.7} & \result{79.4}{0.4} & \result{75.4}{0.4} & \result{68.6}{0.2} \\
\bottomrule
\end{tabular}
\caption{Mean accuracy across ten calibration sets for LLaMA-33B, with standard deviation denoted in subscript.}
\label{tab:results_tasks_llama_33b}
\end{table*}

\begin{table*}[t]
\scriptsize
\centering
\begin{tabular}{llrrrrrrrrrrr}
\toprule
Method & Dataset & \multicolumn{1}{c}{ARC-e} & \multicolumn{1}{c}{ARC-c} & \multicolumn{1}{c}{BoolQ} & \multicolumn{1}{c}{HS} & \multicolumn{1}{c}{LMBD} & \multicolumn{1}{c}{OBQA} & \multicolumn{1}{c}{PIQA} & \multicolumn{1}{c}{RTE} & \multicolumn{1}{c}{SC} & \multicolumn{1}{c}{WG} & \multicolumn{1}{c}{Mean} \\
\midrule
- & - & \multicolumn{1}{c}{74.3} & \multicolumn{1}{c}{42.5} & \multicolumn{1}{c}{78.1} & \multicolumn{1}{c}{56.3} & \multicolumn{1}{c}{62.9} & \multicolumn{1}{c}{34.6} & \multicolumn{1}{c}{77.3} & \multicolumn{1}{c}{68.6} & \multicolumn{1}{c}{79.2} & \multicolumn{1}{c}{69.3} & \multicolumn{1}{c}{64.3} \\
\cmidrule{1-13}
\multirow[c]{5}{*}{GPTQ} & C4 & \result{73.4}{0.3} & \result{41.5}{0.6} & \result{76.4}{1.2} & \result{55.6}{0.2} & \result{60.3}{0.4} & \result{32.3}{0.7} & \result{76.7}{0.4} & \result{66.8}{2.4} & \result{77.8}{0.3} & \result{68.5}{0.4} & \result{62.9}{0.2} \\
 & CNN-DM & \result{73.3}{0.5} & \result{40.7}{0.9} & \result{75.3}{1.0} & \result{55.2}{0.1} & \result{60.1}{0.7} & \result{32.0}{0.7} & \result{76.4}{0.5} & \result{68.7}{1.7} & \result{77.9}{0.6} & \result{67.7}{0.8} & \result{62.7}{0.3} \\
 & RedPajama & \result{73.3}{0.5} & \result{41.4}{0.9} & \result{76.5}{1.0} & \result{55.8}{0.1} & \result{61.5}{0.6} & \result{32.5}{0.6} & \result{76.9}{0.5} & \result{67.0}{1.1} & \result{78.3}{0.5} & \result{68.8}{0.7} & \result{63.2}{0.2} \\
 & RefinedWeb & \result{73.0}{0.5} & \result{41.4}{0.8} & \result{76.6}{0.9} & \result{55.6}{0.2} & \result{61.1}{0.7} & \result{33.1}{0.9} & \result{76.9}{0.2} & \result{66.8}{1.3} & \result{78.1}{0.4} & \result{68.4}{0.4} & \result{63.1}{0.2} \\
 & Wikipedia & \result{73.1}{0.2} & \result{41.6}{0.7} & \result{75.6}{1.5} & \result{55.6}{0.2} & \result{60.5}{0.4} & \result{32.4}{0.8} & \result{77.0}{0.2} & \result{66.1}{1.8} & \result{78.1}{0.5} & \result{68.4}{0.9} & \result{62.9}{0.3} \\
\bottomrule
\end{tabular}
\caption{Mean accuracy across ten calibration sets for Vicuna-7B, with standard deviation denoted in subscript.}
\label{tab:results_tasks_vicuna_7b_v1.3}
\end{table*}

\begin{table*}[t]
\scriptsize
\centering
\begin{tabular}{llrrrrrrrrrrr}
\toprule
Method & Dataset & \multicolumn{1}{c}{ARC-e} & \multicolumn{1}{c}{ARC-c} & \multicolumn{1}{c}{BoolQ} & \multicolumn{1}{c}{HS} & \multicolumn{1}{c}{LMBD} & \multicolumn{1}{c}{OBQA} & \multicolumn{1}{c}{PIQA} & \multicolumn{1}{c}{RTE} & \multicolumn{1}{c}{SC} & \multicolumn{1}{c}{WG} & \multicolumn{1}{c}{Mean} \\
\midrule
- & - & \multicolumn{1}{c}{75.4} & \multicolumn{1}{c}{45.6} & \multicolumn{1}{c}{82.8} & \multicolumn{1}{c}{59.7} & \multicolumn{1}{c}{67.2} & \multicolumn{1}{c}{36.4} & \multicolumn{1}{c}{78.3} & \multicolumn{1}{c}{67.5} & \multicolumn{1}{c}{80.8} & \multicolumn{1}{c}{71.2} & \multicolumn{1}{c}{66.5} \\
\cmidrule{1-13}
\multirow[c]{5}{*}{GPTQ} & C4 & \result{75.3}{0.5} & \result{45.0}{0.7} & \result{82.0}{0.3} & \result{59.0}{0.2} & \result{66.8}{0.4} & \result{35.1}{0.7} & \result{78.2}{0.3} & \result{67.8}{1.4} & \result{80.8}{0.5} & \result{70.7}{0.6} & \result{66.1}{0.2} \\
 & CNN-DM & \result{75.5}{0.4} & \result{45.1}{0.6} & \result{82.6}{0.4} & \result{59.0}{0.2} & \result{66.7}{0.5} & \result{34.8}{0.9} & \result{78.1}{0.5} & \result{67.2}{1.7} & \result{80.9}{0.3} & \result{71.0}{0.4} & \result{66.1}{0.2} \\
 & RedPajama & \result{75.6}{0.3} & \result{45.3}{0.4} & \result{82.3}{0.4} & \result{59.1}{0.2} & \result{66.7}{0.5} & \result{34.8}{0.5} & \result{78.2}{0.4} & \result{68.5}{1.6} & \result{80.8}{0.4} & \result{70.2}{0.4} & \result{66.2}{0.2} \\
 & RefinedWeb & \result{75.7}{0.2} & \result{45.3}{0.8} & \result{82.1}{0.3} & \result{59.0}{0.1} & \result{66.4}{0.3} & \result{34.4}{0.8} & \result{78.3}{0.3} & \result{67.3}{2.0} & \result{80.7}{0.5} & \result{70.8}{0.6} & \result{66.0}{0.2} \\
 & Wikipedia & \result{75.6}{0.5} & \result{45.0}{0.5} & \result{82.0}{0.5} & \result{59.0}{0.2} & \result{66.3}{0.5} & \result{34.5}{0.5} & \result{78.2}{0.3} & \result{68.2}{1.2} & \result{80.8}{0.4} & \result{70.2}{0.6} & \result{66.0}{0.1} \\
\bottomrule
\end{tabular}
\caption{Mean accuracy across ten calibration sets for Vicuna-13B, with standard deviation denoted in subscript.}
\label{tab:results_tasks_vicuna_13b_v1.3}
\end{table*}

\begin{table*}[t]
\scriptsize
\centering
\begin{tabular}{llrrrrrrrrrrr}
\toprule
Method & Dataset & \multicolumn{1}{c}{ARC-e} & \multicolumn{1}{c}{ARC-c} & \multicolumn{1}{c}{BoolQ} & \multicolumn{1}{c}{HS} & \multicolumn{1}{c}{LMBD} & \multicolumn{1}{c}{OBQA} & \multicolumn{1}{c}{PIQA} & \multicolumn{1}{c}{RTE} & \multicolumn{1}{c}{SC} & \multicolumn{1}{c}{WG} & \multicolumn{1}{c}{Mean} \\
\midrule
- & - & \multicolumn{1}{c}{78.5} & \multicolumn{1}{c}{50.3} & \multicolumn{1}{c}{83.9} & \multicolumn{1}{c}{61.9} & \multicolumn{1}{c}{70.7} & \multicolumn{1}{c}{34.2} & \multicolumn{1}{c}{79.2} & \multicolumn{1}{c}{74.0} & \multicolumn{1}{c}{81.2} & \multicolumn{1}{c}{74.6} & \multicolumn{1}{c}{68.8} \\
\cmidrule{1-13}
\multirow[c]{5}{*}{GPTQ} & C4 & \result{77.8}{0.3} & \result{49.5}{0.5} & \result{83.7}{0.4} & \result{61.3}{0.2} & \result{69.9}{0.4} & \result{35.2}{1.0} & \result{78.7}{0.2} & \result{73.0}{1.7} & \result{81.1}{0.3} & \result{73.0}{0.5} & \result{68.3}{0.2} \\
 & CNN-DM & \result{77.5}{0.5} & \result{49.2}{0.9} & \result{83.2}{0.5} & \result{61.2}{0.1} & \result{69.4}{0.4} & \result{35.7}{0.7} & \result{78.8}{0.4} & \result{73.4}{1.2} & \result{81.5}{0.2} & \result{73.7}{0.7} & \result{68.4}{0.2} \\
 & RedPajama & \result{77.9}{0.3} & \result{49.7}{0.4} & \result{84.0}{0.4} & \result{61.3}{0.2} & \result{69.9}{0.4} & \result{35.4}{0.4} & \result{78.8}{0.4} & \result{73.0}{1.4} & \result{81.2}{0.2} & \result{73.3}{0.6} & \result{68.4}{0.2} \\
 & RefinedWeb & \result{77.8}{0.3} & \result{49.7}{0.4} & \result{83.7}{0.4} & \result{61.3}{0.1} & \result{69.6}{0.4} & \result{35.4}{0.8} & \result{78.9}{0.3} & \result{73.0}{1.1} & \result{80.9}{0.3} & \result{73.6}{0.5} & \result{68.4}{0.2} \\
 & Wikipedia & \result{77.5}{0.5} & \result{48.8}{0.6} & \result{83.1}{0.3} & \result{61.2}{0.1} & \result{69.4}{0.4} & \result{35.3}{0.8} & \result{78.8}{0.2} & \result{72.9}{1.7} & \result{81.1}{0.4} & \result{73.4}{0.4} & \result{68.1}{0.2} \\
\bottomrule
\end{tabular}
\caption{Mean accuracy across ten calibration sets for Vicuna-33B, with standard deviation denoted in subscript.}
\label{tab:results_tasks_vicuna_33b_v1.3}
\end{table*}

\begin{table*}[t]
\scriptsize
\centering
\begin{tabular}{llrrrrrrrrrrr}
\toprule
Method & Dataset & \multicolumn{1}{c}{ARC-e} & \multicolumn{1}{c}{ARC-c} & \multicolumn{1}{c}{BoolQ} & \multicolumn{1}{c}{HS} & \multicolumn{1}{c}{LMBD} & \multicolumn{1}{c}{OBQA} & \multicolumn{1}{c}{PIQA} & \multicolumn{1}{c}{RTE} & \multicolumn{1}{c}{SC} & \multicolumn{1}{c}{WG} & \multicolumn{1}{c}{Mean} \\
\midrule
- & - & \multicolumn{1}{c}{65.6} & \multicolumn{1}{c}{30.5} & \multicolumn{1}{c}{66.1} & \multicolumn{1}{c}{50.5} & \multicolumn{1}{c}{63.3} & \multicolumn{1}{c}{27.6} & \multicolumn{1}{c}{76.3} & \multicolumn{1}{c}{55.2} & \multicolumn{1}{c}{73.6} & \multicolumn{1}{c}{65.2} & \multicolumn{1}{c}{57.4} \\
\cmidrule{1-13}
\multirow[c]{5}{*}{GPTQ} & C4 & \result{65.2}{0.4} & \result{30.9}{0.5} & \result{63.6}{2.6} & \result{49.9}{0.1} & \result{60.3}{1.6} & \result{26.8}{0.6} & \result{76.1}{0.3} & \result{54.5}{1.2} & \result{73.3}{0.3} & \result{65.1}{0.4} & \result{56.6}{0.4} \\
 & CNN-DM & \result{65.0}{0.4} & \result{30.9}{0.5} & \result{63.5}{1.9} & \result{49.9}{0.1} & \result{60.7}{1.1} & \result{26.6}{0.7} & \result{76.1}{0.3} & \result{55.1}{1.6} & \result{73.4}{0.3} & \result{64.3}{0.7} & \result{56.5}{0.3} \\
 & RedPajama & \result{65.1}{0.3} & \result{30.9}{0.6} & \result{63.7}{1.8} & \result{49.9}{0.1} & \result{62.4}{0.9} & \result{26.3}{0.9} & \result{76.0}{0.3} & \result{54.9}{1.2} & \result{73.2}{0.2} & \result{64.4}{0.6} & \result{56.7}{0.3} \\
 & RefinedWeb & \result{65.1}{0.5} & \result{31.1}{0.7} & \result{63.3}{2.2} & \result{49.9}{0.1} & \result{60.8}{1.7} & \result{26.4}{0.9} & \result{76.1}{0.2} & \result{55.2}{0.9} & \result{73.1}{0.5} & \result{63.9}{0.7} & \result{56.5}{0.4} \\
 & Wikipedia & \result{65.1}{0.4} & \result{31.0}{0.3} & \result{64.4}{1.6} & \result{49.8}{0.1} & \result{61.0}{1.4} & \result{26.7}{0.7} & \result{76.1}{0.3} & \result{55.3}{1.4} & \result{73.2}{0.3} & \result{64.7}{0.6} & \result{56.7}{0.3} \\
\bottomrule
\end{tabular}
\caption{Mean accuracy across ten calibration sets for OPT-6.7B, with standard deviation denoted in subscript.}
\label{tab:results_tasks_opt_6.7b}
\end{table*}

\begin{table*}[t]
\scriptsize
\centering
\begin{tabular}{llrrrrrrrrrrr}
\toprule
Method & Dataset & \multicolumn{1}{c}{ARC-e} & \multicolumn{1}{c}{ARC-c} & \multicolumn{1}{c}{BoolQ} & \multicolumn{1}{c}{HS} & \multicolumn{1}{c}{LMBD} & \multicolumn{1}{c}{OBQA} & \multicolumn{1}{c}{PIQA} & \multicolumn{1}{c}{RTE} & \multicolumn{1}{c}{SC} & \multicolumn{1}{c}{WG} & \multicolumn{1}{c}{Mean} \\
\midrule
- & - & \multicolumn{1}{c}{67.1} & \multicolumn{1}{c}{32.9} & \multicolumn{1}{c}{65.9} & \multicolumn{1}{c}{52.4} & \multicolumn{1}{c}{63.1} & \multicolumn{1}{c}{27.2} & \multicolumn{1}{c}{75.8} & \multicolumn{1}{c}{57.8} & \multicolumn{1}{c}{76.0} & \multicolumn{1}{c}{65.0} & \multicolumn{1}{c}{58.3} \\
\cmidrule{1-13}
\multirow[c]{5}{*}{GPTQ} & C4 & \result{66.9}{0.3} & \result{33.1}{0.3} & \result{64.9}{2.1} & \result{51.6}{0.1} & \result{61.4}{0.8} & \result{26.4}{0.6} & \result{75.9}{0.3} & \result{56.9}{1.8} & \result{75.0}{0.4} & \result{65.2}{0.5} & \result{57.7}{0.3} \\
 & CNN-DM & \result{67.1}{0.4} & \result{33.0}{0.5} & \result{65.7}{1.4} & \result{51.6}{0.1} & \result{61.6}{0.6} & \result{26.7}{1.1} & \result{76.0}{0.3} & \result{56.2}{1.7} & \result{75.1}{0.4} & \result{64.8}{0.6} & \result{57.8}{0.3} \\
 & RedPajama & \result{67.1}{0.6} & \result{33.2}{0.5} & \result{65.5}{1.6} & \result{51.6}{0.1} & \result{61.9}{0.6} & \result{26.8}{0.6} & \result{75.7}{0.2} & \result{56.4}{1.1} & \result{75.3}{0.2} & \result{65.4}{0.3} & \result{57.9}{0.3} \\
 & RefinedWeb & \result{66.7}{0.4} & \result{32.8}{0.4} & \result{66.0}{1.6} & \result{51.6}{0.1} & \result{61.6}{0.5} & \result{26.5}{0.3} & \result{75.8}{0.4} & \result{57.0}{1.3} & \result{75.2}{0.2} & \result{64.9}{0.5} & \result{57.8}{0.2} \\
 & Wikipedia & \result{67.2}{0.4} & \result{32.9}{0.3} & \result{65.2}{1.6} & \result{51.6}{0.1} & \result{61.9}{0.6} & \result{26.7}{0.9} & \result{75.7}{0.2} & \result{57.9}{1.4} & \result{75.2}{0.2} & \result{64.3}{0.6} & \result{57.9}{0.2} \\
\bottomrule
\end{tabular}
\caption{Mean accuracy across ten calibration sets for OPT-13B, with standard deviation denoted in subscript.}
\label{tab:results_tasks_opt_13b}
\end{table*}

\begin{table*}[t]
\scriptsize
\centering
\begin{tabular}{llrrrrrrrrrrr}
\toprule
Method & Dataset & \multicolumn{1}{c}{ARC-e} & \multicolumn{1}{c}{ARC-c} & \multicolumn{1}{c}{BoolQ} & \multicolumn{1}{c}{HS} & \multicolumn{1}{c}{LMBD} & \multicolumn{1}{c}{OBQA} & \multicolumn{1}{c}{PIQA} & \multicolumn{1}{c}{RTE} & \multicolumn{1}{c}{SC} & \multicolumn{1}{c}{WG} & \multicolumn{1}{c}{Mean} \\
\midrule
- & - & \multicolumn{1}{c}{70.0} & \multicolumn{1}{c}{34.6} & \multicolumn{1}{c}{70.5} & \multicolumn{1}{c}{54.3} & \multicolumn{1}{c}{64.3} & \multicolumn{1}{c}{30.2} & \multicolumn{1}{c}{77.6} & \multicolumn{1}{c}{57.4} & \multicolumn{1}{c}{76.6} & \multicolumn{1}{c}{68.4} & \multicolumn{1}{c}{60.4} \\
\cmidrule{1-13}
\multirow[c]{5}{*}{GPTQ} & C4 & \result{69.5}{0.3} & \result{34.0}{0.6} & \result{69.7}{0.7} & \result{53.7}{0.1} & \result{62.8}{0.8} & \result{29.2}{0.5} & \result{77.4}{0.2} & \result{55.9}{1.8} & \result{76.3}{0.2} & \result{68.0}{0.4} & \result{59.6}{0.2} \\
 & CNN-DM & \result{69.4}{0.2} & \result{33.7}{0.4} & \result{69.4}{0.7} & \result{53.6}{0.1} & \result{63.3}{0.8} & \result{29.4}{0.6} & \result{77.4}{0.2} & \result{55.8}{1.3} & \result{76.4}{0.3} & \result{67.9}{0.3} & \result{59.6}{0.1} \\
 & RedPajama & \result{69.5}{0.4} & \result{33.4}{0.4} & \result{70.1}{0.3} & \result{53.6}{0.1} & \result{63.5}{0.4} & \result{29.5}{0.5} & \result{77.4}{0.3} & \result{56.4}{2.1} & \result{76.3}{0.2} & \result{67.7}{0.3} & \result{59.8}{0.3} \\
 & RefinedWeb & \result{69.5}{0.4} & \result{33.5}{0.4} & \result{69.8}{1.1} & \result{53.7}{0.1} & \result{63.1}{0.6} & \result{29.3}{0.4} & \result{77.3}{0.2} & \result{56.1}{1.5} & \result{76.3}{0.2} & \result{67.9}{0.9} & \result{59.7}{0.2} \\
 & Wikipedia & \result{69.5}{0.3} & \result{33.5}{0.3} & \result{69.6}{0.7} & \result{53.6}{0.1} & \result{62.8}{0.8} & \result{29.6}{0.4} & \result{77.5}{0.2} & \result{56.5}{1.1} & \result{76.4}{0.3} & \result{67.9}{0.4} & \result{59.7}{0.2} \\
\bottomrule
\end{tabular}
\caption{Mean accuracy across ten calibration sets for OPT-30B, with standard deviation denoted in subscript.}
\label{tab:results_tasks_opt_30b}
\end{table*}

\end{document}